\pdfoutput=1

\documentclass[11pt]{article}

\usepackage[preprint]{acl}

\usepackage{times}
\usepackage{latexsym}

\usepackage[T1]{fontenc}

\usepackage[utf8]{inputenc}

\usepackage{microtype}

\usepackage{inconsolata}

\usepackage{graphicx}

\usepackage{algorithm}
\usepackage{algorithmic}

\usepackage{multirow}
\usepackage{booktabs}
\usepackage{tabularx}
\usepackage{multirow}
\usepackage{float}
\usepackage{todonotes}

\usepackage{multirow}
\usepackage{tabularx}
\usepackage{booktabs}
\usepackage{xcolor,caption,subcaption,graphicx}
\usepackage{amssymb}
\usepackage{amsmath}
\usepackage{amsthm}

\title{Skim-Aware Contrastive Learning for Efficient Document Representation}

\author{Waheed Ahmed Abro\\
  CDEP - UR 2471, Univ Artois, France\\
  \texttt{wahmed.abro@univ-artois.fr} \\\And
  Zied Bouraoui \\
  CRIL - CNRS, Univ Artois,  France \\
  \texttt{zied.bouraoui@cril.fr} \\}

\begin{document}
\maketitle
\begin{abstract}
Although transformer-based models have shown strong performance in word- and sentence-level tasks, effectively representing long documents, especially in fields like law and medicine, remains difficult. Sparse attention mechanisms can handle longer inputs, but are resource-intensive and often fail to capture full-document context. Hierarchical transformer models offer better efficiency but do not clearly explain how they relate different sections of a document. In contrast, humans often skim texts, focusing on important sections to understand the overall message. Drawing from this human strategy, we introduce a new self-supervised contrastive learning framework that enhances long document representation. Our method randomly masks a section of the document and uses a natural language inference (NLI)-based contrastive objective to align it with relevant parts while distancing it from unrelated ones. This mimics how humans synthesize information, resulting in representations that are both richer and more computationally efficient. Experiments on legal and biomedical texts confirm significant gains in both accuracy and efficiency. 
\end{abstract}

\section{Introduction}

Since the introduction of Language Models (LMs), the focus in NLP has been on fine-tuning large pre-trained language models, especially for solving sentence and paragraph-level tasks. However, accurately learning document embeddings continues to be an important challenge for several applications, such as document classification~\cite{saggau2023efficient},  ranking~\cite{ ginzburg2021self, izacard2021unsupervised}, retrieval-augmented generation (RAG) systems that demand efficient document representation encoders~\cite{zhang2024mgte, zhao2025funnelrag}
and legal and medical applications like judgment prediction~\cite{chalkidis2019neural, feng2022legal}, legal information retrieval~\cite{ SANSONE2022LIR}, and biomedical document classification~\cite{Johnson2016MIMICIIIAF,wang2023pre}. 

Learning high-quality document representations is a challenging task due to the difficulty in developing efficient encoders with reasonable complexity. Most document encoders use sentence encoders based on self-attention architectures such as BERT \cite{devlin2019bert}. However, it is not feasible to have inputs that are too long as self-attention scales quadratically with the input length. 
To process long inputs efficiently, architectures such as Linformer \cite{DBLP:journals/corr/abs-2006-04768}, Big Bird \cite{DBLP:conf/nips/ZaheerGDAAOPRWY20}, Longformer \cite{DBLP:journals/corr/abs-2004-05150} and Hierarchical Transformers \cite{DBLP:journals/corr/abs-2210-05529} have been developed. Unlike quadratic scaling in traditional attention mechanisms, these architectures utilize sparse attention mechanisms or hierarchical attention mechanisms that scale linearly. As such, they can process $4096$ input tokens, which is enough to embed most types of documents, including legal and medical documents, among others. 

While methods based on sparse attention networks offer a solution for complexity, the length of the document remains a problem for building faithful representations for downstream applications such as legal and medical domains. First, fine-tuning these models for downstream tasks is computationally intensive. Second, capturing the meaning of the whole document remains too complex. In particular, it is unclear how or to what extent inner-dependencies between text fragments are considered. 
This is because longer documents contain more information than shorter documents, making it difficult to capture all the relevant information within a fixed-size representation. Additionally, documents usually cover different parts, making the encoding process complex and may lead to collapsed representations. 
This is particularly true for legal and medical documents, as they contain specialized terminology and text segments that describe a series of interrelated facts. 

When domain experts, such as legal or medical professionals, read through documents, they skillfully skim the text, honing in on some text fragments that, when pieced together, provide an understanding of the content. Inspired by this intuitive process, our work focuses on developing document encoders capable of generating high-quality embeddings for long documents. These encoders mimic the expert ability to distill relevant text chunks, enabling them to excel in downstream tasks right out of the box, without the need for fine-tuning. We propose a novel framework for self-supervised contrastive learning that focuses on long legal and medical documents. Our approach features a self-supervised Chunk Prediction Encoder (CPE) designed to tackle the challenge of learning document representations without supervision. By leveraging both intra-document and inter-document chunk relationships, the CPE enhances how documents should be represented through its important fragments. The model operates by randomly selecting a text fragment from a document and using an encoder to predict whether this fragment is strongly related to other parts of the same document. To simulate the skimming process, we frame this task as a Natural Language Inference (NLI) problem. In our method, “entailment” and “contradiction” are not literal NLI labels. Rather, we use an NLI-style binary objective as a practical proxy that teaches the model whether a chunk is semantically compatible with its document context (positive) or not (negative). This enables the encoder to judge whether a local fragment is compatible with document-level context, thereby capturing long-range, cross-fragment relevance. This method not only uncovers connections between different documents, but also emphasizes the relevance of various sections, thus enriching the overall learning process for document representations.
The main contributions are as follows:
\begin{itemize}
    \item We introduce a self-supervised Chunk Prediction Encoder (CPE) that employs random span sampling into a hierarchical transformer and Longformer: by sampling text spans and training the model to predict whether a span belongs to the same document, CPE captures both intra- and inter-document fragment relationships, preserves global context through chunk aggregation, and models the complex hierarchical structures of long texts.

    \item We apply a contrastive loss that pulls together representations of related fragments and pushes apart unrelated ones, reinforcing meaningful connections across different parts of the same document.
    
    \item We conducted extensive experiments to demonstrate the effectiveness of our framework. Specifically, i) we compare the quality of our document embeddings against strong baselines, 
    (ii) we benchmark CPE-based models in an end-to-end fine-tuning setup 
    while 
    training all parameters jointly and comparing downstream classification performance against established methods, and (iii) we perform an ablation study to assess the impact of different chunk sizes, visualize the resulting embedding space, and evaluate performance on shorter documents. We also compare with LLMs such as LLaMA  capabilities for handling long legal texts.

\end{itemize}

\section{Related Work} \label{sec:RelWork}

This section provides an overview of the modelling of long documents and self-supervised document embedding. 

\paragraph{Modeling of long documents.} 
Long documents are typically handled using sparse-attention models such as Longformer~\cite{Beltagy2020Longformer} and BigBird~\cite{Zaheer2020BigBird}. These models use local and global attention mechanisms to overcome the $O(n^2)$ complexity of standard full attention mechanisms. Alternatively, one can use a hierarchical attention mechanism~\cite{yang2016hierarchical}, where the document is processed in a hierarchical manner. For example, \cite{chalkidis2019neural} applied a hierarchical BERT model to model long legal documents. The model first processes the words in each sentence using the BERT-base model to produce sentence embeddings. Then, a self-attention mechanism is applied to the sentence-level embeddings to produce document embeddings. The authors have demonstrated that their hierarchical BERT model outperforms both the Vanilla BERT architecture and Bi-GRU models. Similarly,~\cite{dai2022revisiting} explored the various methods of splitting the long document and compared them with sparse attention methods on long document classification tasks.  Their findings showed that better results were achieved by splitting the document into smaller chunks of 128 tokens. ~\cite{wu2021hi}  proposed a model called Hi-Transformer, which applies both sentence-level and document-level transformers followed by hierarchical pooling. Meanwhile, \cite{chalkidis2022exploration} introduced a variant of hierarchical attention transformers based on segment encoder and cross-segment encoder, which demonstrated comparable results with the Longformer model. In this work, we consider processing long documents with a hierarchical attention mechanism and sparse-attention Longformer encoders. We further improve document embedding using self-supervised contrastive learning.

\paragraph{Unsupervised Document Representation.}
Unsupervised document representation learning is a highly active research field. At first, deep learning models were introduced to create contextualized word representations, such as Word2Vec~\cite{word2vec2013} and GloVE~\cite{pennington2014glove}. The Doc2Vec~\cite{pmlr2014doc2vec} model was proposed, which utilized contextualized word representation to generate document embeddings. In the same vein, the Skip-Thoughts~\cite{Skip-Thought2015} model extended the word2vec approach from the word level to the sentence level. Transformer-based models~\cite{Reimers2019SBERT} were also suggested to produce a vector representation of the sentence. 
Recently, there have been advancements in self-supervised contrastive learning methods ~\cite{iter2020pretraining,gao2021simcse,giorgi2021declutr,klein2022scd,saggau2023efficient}. 
In this direction, CONPONO proposes using sentence-level objectives with a masked language model to capture discourse coherence between sentences. The sentence-level encoder predicts text that is k sentences away.
On the other hand, SimCSE~\cite{gao2021simcse} uses a dropout version of the same sentence as a positive pair on short sentences. Similarly, \cite{saggau2023efficient} proposed SimCSE learning on long documents with additional Bregman divergence. On the other hand, SDDE~\cite{chen2019self} model was proposed to generate document representation based on inter-document relationships using an RNN encoder. We follow a similar strategy of exploiting inter-document relationships, we employe transformer-based pre-trained language models with multiple negatives ranking contrastive loss.  
   

\paragraph{Legal and medical document representation}
Processing legal and medical documents is an active research topic. \cite{chalkidis2019neural} propose the hierarchical BERT model to process legal documents. \cite{malik2021ildc} propose the hierarchical transformer model architecture for the legal judgment prediction task. The input document is split into several chunks of size 512 tokens. Each chunk embedding is produced by a pre-trained XLNET model. Then, a Bi-GRU encoder is applied to the chunk embeddings to produce final document embeddings. \cite{zheng2021does} train the BERT model on CaseHOLD
(Case Holdings On Legal Decisions) dataset. \cite{hamilton2023blind} employed GPT-2 models to predict how each justice votes for supreme court justice’s opinions. \cite{ijcai2019Multi} process the legal document using a Multi-Perspective Bi-Feedback Network to classify law articles. \cite{xu2020distinguish} propose to represent the document using a graph neural network. To distinguish confusing articles, a distillation-based attention network is applied to extract 
discriminative features.  

For medical document processing, in \cite{arnold2020learning}, authors propose contextualized document representations to answer questions from long medical documents. The model employs a hierarchical dual encoder based on hierarchical LSTM to encode medical entities. \cite{mullenbach2018explainable} proposed a convolutional neural network base label-wise attention network to produce label-wise document representations by attending to the most relevant document information for each medical code. In \cite{ChalkidisFKMAA20}, authors showed that the pre-trained BERT model outperforms CNN-based label-wise attention networks. In the same direction, \cite{kementchedjhieva2023exploration} proposed encoder-decoder architecture and outperforms the encoder-only model on multi-label text classification for legal and medical domains. The work in \cite{wang2023pre} provides a comprehensive survey focusing on the integration of pre-trained language models within the biomedical domain. Their investigation highlights the substantial benefits of employing LMs in various NLP tasks. In contrast to previous works, we propose the learn document representation using self-supervised contrastive learning pre-trained LMs.  

\begin{figure}[t!]
\includegraphics[width=0.45\textwidth]{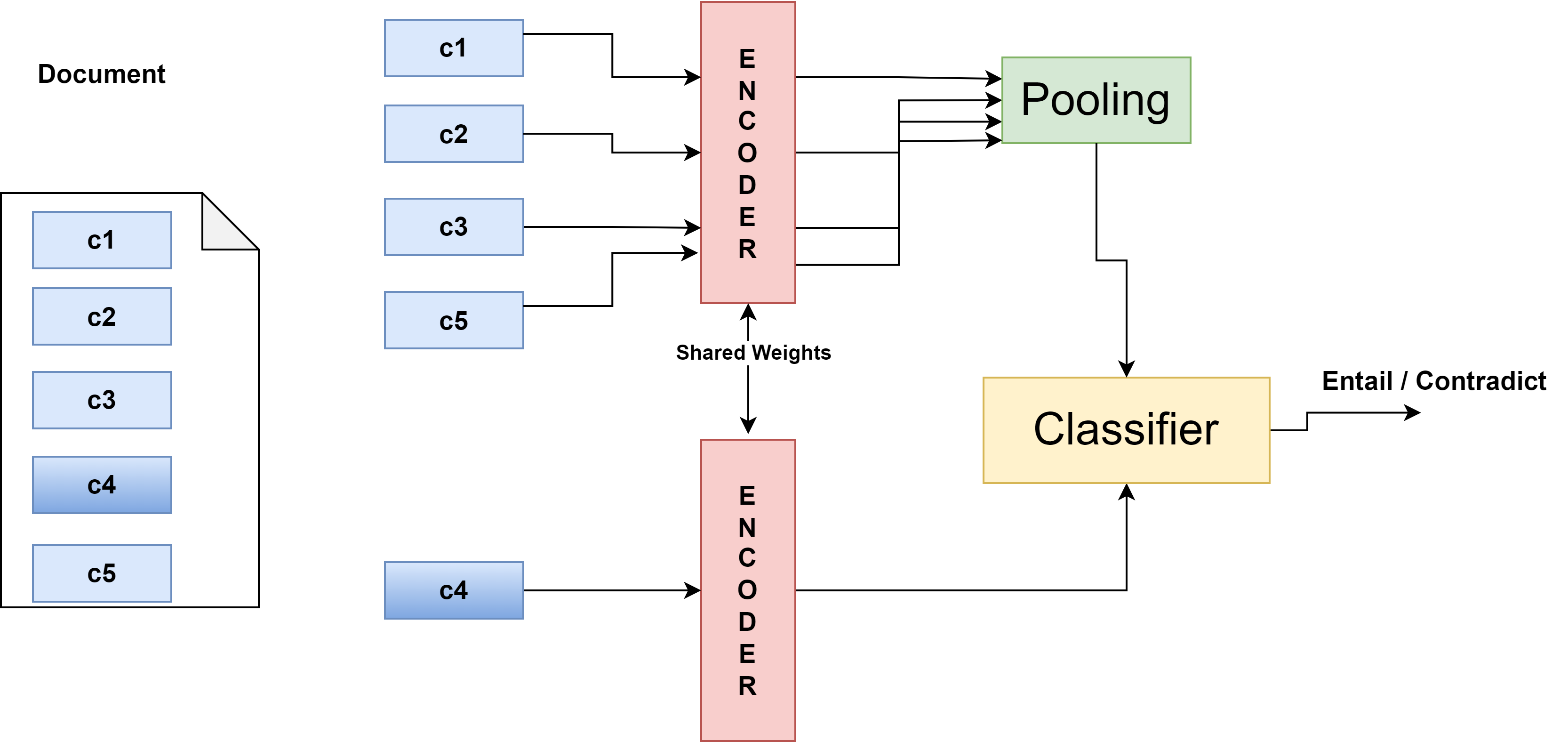}
\caption{Illustration of CPE Contrastive Learning via the hierarchical transformer model.}
\label{fig_HiCPE}
\end{figure}  

\section{Chunk Prediction Encoders} \label{sec:CPE}
To process long documents in time efficient manner, we use state-of-the-art hierarchical transformer method ~\cite{chen2019self,wu2021hi,dai2022revisiting} and sparse attention longformer model~\cite{Beltagy2020Longformer}. To enhance the quality of these representations, we propose self-supervised CPE that takes into account the relationship between different text chunks and determines relevant ones. Finally, we use the document embedding to predict the outcome of single and multi-label classification for legal and medical classification tasks.

\paragraph{CPE for Hierarchical Representation.}
\label{doc_representation}
We first briefly introduce hierarchical representation model using pre-trained model $\mathcal{M}$. Let $D$ be an input document, and $c_1, c_2, ..., c_n$ denotes the set of corresponding text chunks in $D$ where $n$ is the maximum number of chunks, padding with zero if the chunks are less than $n$. Each chunk contains a sequence of $T$ tokens $C = (w_{1}, w_{2},.., w_{t}) $, where $t$ is less than $512$. Furthermore, the special classification \verb|[CLS]| token is added at the start of each chunk. Our aim is to learn the vector representation of each chunk using a shared \textit{small} language model encoder as follows:
\begin{equation}
\label{eq_sent_rep} 
f = \mathcal{M}(w_{\text{[CLS]}}, w_{1}, \ldots, w_{t})
\end{equation}
where $f$ represents the output of a model (which can be BERT, RoBERTa, LegalBERT, ClinicalBioBERT or any other). Following the common strategy, we consider the \verb|[CLS]| token as the representation of the whole chunk.  To obtain the final document representation from different chunk features, we consider the following pooling strategies: the Mean-Pooling obtained by taking the mean of chunks representation $d_t = 1/ n \sum_{i=1}^n {f_i}$, and the Max-Pooling over chunks representation. Each chunk encodes the local feature of the document, and the whole document is represented by the average of these local features.

\begin{figure}[t!]
\includegraphics[width=0.45\textwidth]{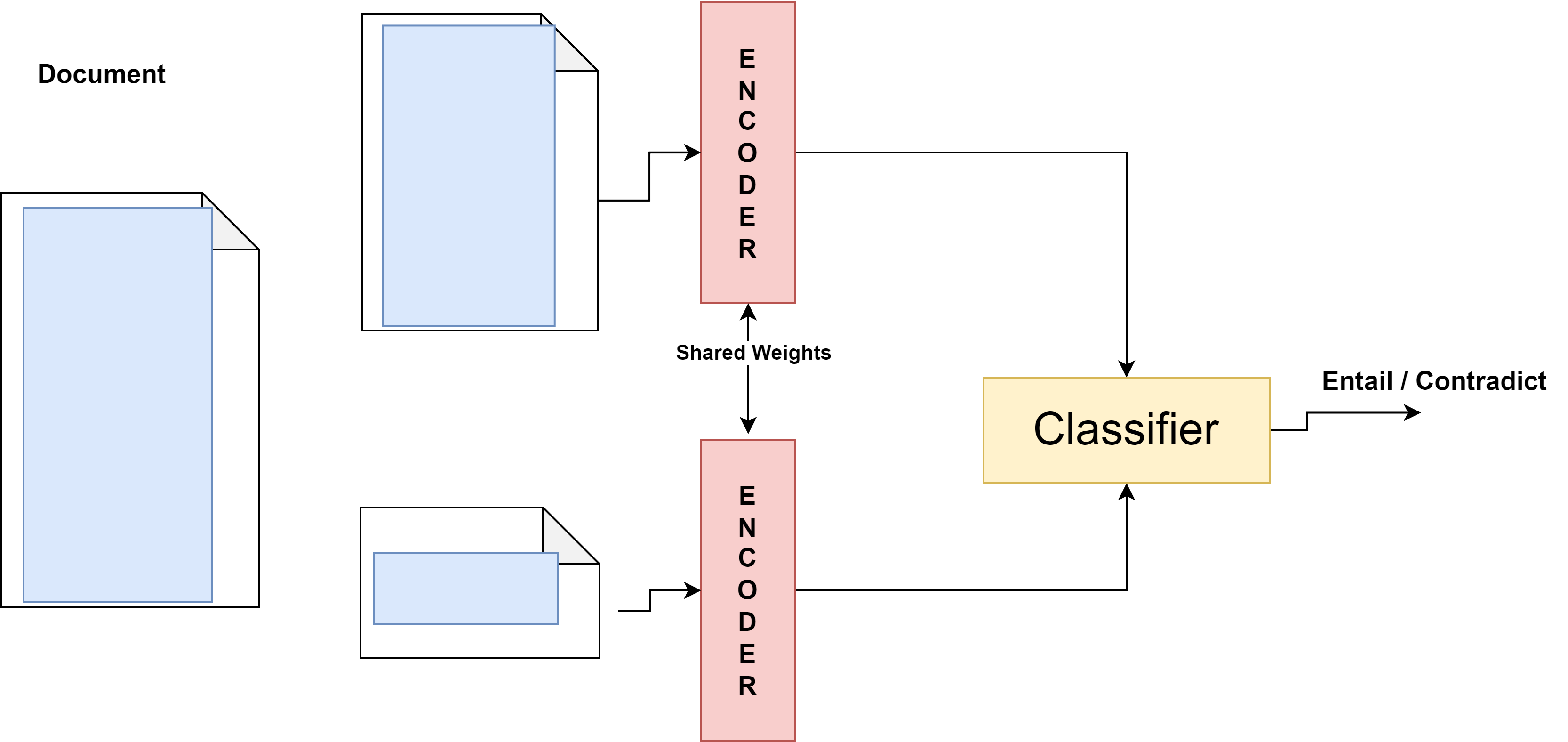}
\caption{The CPE Contrastive Learning process with the Longformer model.}
\label{fig_LongCPE}
\end{figure}

\paragraph{Learning process}
\label{CPE_contrastive}
We propose a chunk prediction encoder to train a hierarchical transformer model using self-supervised contrastive learning to leverage intra and inter-document relationships. 
For each document, we randomly remove one chunk and then ask the NLI classifier to predict whether this chunk is derived from other chunks of the document. By doing so, we force the model to learn the dependencies between chunks and their relevance in representing a document. Consider a mini-batch of $N$ documents, denoted as $D=\left\{\left( d_i\right )\right\}_1 ^{N}$. For each document ${d_i}$, we randomly select a text chunk $c^+$ and remove it from that document $\tilde{d_i}$ to form a positive pair $\left ( \tilde{d_i}, c^+  \right )$. We then select a negative chunk $c^-$ from the remaining $N-1$ documents of the batch to form a negative pair, $(\tilde{d_i}, c^-)$. Notice that $c^-$ does not belong to document $\tilde{d_i}$. One concern that could be seen here is the text chunks could be similar and fit on both positive and negative documents. However, this is not an issue as in the training objective, we have multiple negatives, so our model is forced to optimize most dissimilar documents than most similar ones. The chunk predictive contrastive learning process can be viewed as an unsupervised natural language inference task, where a positive chunk sample represents an entailment of a document, and negative samples from other documents represent a contradiction of the document. We use multiple negatives ranking loss~\cite{henderson2017efficient} to train the model:
\begin{equation}
    \mathcal{L} = -\frac{1}{n}\sum_{i=1}^{n}  \frac{\exp( sim (f(\tilde{d_i}), f(c_i^+)))}{\sum_{1}^{k} \exp(sim (f (\tilde{d_i}), f(c_k^-)))},
    \label{eq_loss}
\end{equation}
where $f$ denotes the feature vector generated by document encoder, \textit{sim} represents cosine similarity, $c_i^+$ is the positive chunk taken from the document $d_i$ and $c_k^-$ are the $k$ negative sample of chunk taken from other documents than $d_i$. Multiple negative ranking loss compares the positive pair representation with the negative pair samples in mini-batch.  The general architecture is illustrated in Figure~\ref{fig_HiCPE}. Our architecture includes a shared encoder that generates a complete document embedding, except for chunk $C4$. Additionally, the shared encoder shown at the bottom produces a vector representation of $C4$. The classifier is responsible for learning whether the embedding of chunk $C4$ aligns with or contradicts the document embedding.

\paragraph{CPE for Longformer} \label{lab:Longformer}

During the training of the CPE on the Longformer model, the document is split into two segments: the reference text and the positive text chunk. Our approach aims to ensure that the reference text segment and the positively sampled text chunk, which is extracted from the same document, are in agreement and maximize in consequence the objective function.  
Suppose we have a mini-batch of $N$ documents, denoted as $D=\left\{\left( d_i\right )\right\}_1 ^{N}$. For each document ${d_i}$, we randomly select a text chunk $c^+$ and the remaining text as reference text $\tilde{d}$ to form a positive pair $\left (\tilde{d_i}, c_i^+  \right )$. We then select a negative chunk $c^-$ from the remaining $N-1$ documents of the batch to serve as the negative pair, $(\tilde{d_i}, c_j^-)$. The reference text and positive text chunks are passed through the same Longformer encoder. The proposed method utilizes the [CLS] token representations to produce the reference text ($z_{\tilde{d_i}}$) embedding and text segment ($z_c^+$) embedding. The linear classifier is then trained to maximize the agreement between the segment and reference text, both sampled from the same document and minimize the agreement between reference text and segment taken from another document.  The multiple negatives ranking loss is used to optimize the model as given as follows: 
\begin{equation}
    \mathcal{L} = -\frac{1}{n}\sum_{i=1}^{n}  \frac{\exp(sim (z_{\tilde{d_i}}, z_c^+))}{\sum_{1}^{k} \exp(sim (z_{\tilde{d_i}}, z_{c_k}^-))},
    \label{eq_loss}
\end{equation}

where $z_{\tilde{d_i}}$, $z_{c_i}^+$, and $z_{c_k}^-$ denote the feature vectors generated by the Longformer document encoder.
The general architecture is depicted in Figure~\ref{fig_LongCPE}. 
 A shared Longformer encoder creates a document embedding for most of the text, excluding a small portion that has been removed. Additionally, a shared encoder at the bottom generates a vector representation of the removed chunk. The classifier learns to differentiate whether the embedding of the small chunk aligns with or contradicts the document embedding.

\paragraph{Document Classification}
\label{doc_classif}
After producing our document representations, we train an
MLP classifier, which consists of three hidden layers and an output layer. The hidden layer uses the TANH activation function. In the output layer, we use the sigmoid for the multi-label classifier and the softmax for the multi-class classifier. 
\begin{equation}
\hat{y} = g(tanh(W * d + b))
\label{eq_final}
\end{equation}
where $g$ denotes the activation function (sigmoid or softmax), $d$ is the document representation, $W$ represents the weight matrices, $b$ is the bias vector. The model minimizes the cross-entropy loss. Notice that we introduce this classifier to assess the quality of our document embeddings and to evaluate to what extent the representation can be considered to perform classification tasks.  

\section{Experimental Setting} 
\label{sec:ExpSet}

In this section, we define the experimental setup for evaluating the proposed contrastive learning framework. 

\paragraph{Datasets}

We evaluate our encoders on three legal datasets, ECHR, SCOTUS, and EURLEX and two medical datasets, MIMIC and Biosq. Table~\ref{table-Datastats} in Appendix provides statistics and the average document length of each dataset. Notice that we also consider Biosq to test the capability of our model in handling short-length documents as well.

\paragraph{Setting and evaluation metrics}
We consider small language models such bert-base\footnote{\url{https://huggingface.co/bert-base-uncased}}, roberta-base\footnote{\url{https://huggingface.co/FacebookAI/roberta-base}}, legal-bert-base\footnote{\url{https://huggingface.co/nlpaueb/legal-bert-base-uncased}}, and ClinicalBioBERT\footnote{\url{https://huggingface.co/Dinithi/ClinicalBioBERT}} models as we seek for a simple model with less number of parameters. While our hierarchical transformer is capable of processing documents of any length, in this study, we restrict our analysis to the initial 4096 tokens driven primarily by computational considerations. The number of text chunks is set to 32 and the length of the chunk is set to 128 for ECHR, SCOTUS, and MIMIC datasets. For the EURLEX dataset, which is smaller than other datasets, we utilize initial 2048 tokens; 16 segments of 128 tokens each. The Longformer model can handle sequences of up to 4096 input length. We truncate or pad documents that are longer or shorter accordingly. The length of the positive text chunk is set to 128 tokens for the Longformer model. As evaluation metrics, we report results in Micro-averaged F1 and Macro-averaged F1 for Multi-label ECHR and Multi-class SCOTUS classification datasets.

\begin{table*}[t]
\centering
\caption{Classification Model performance on document embedding produced by contrastively trained pre-trained hierarchical transformer models on ECHR, SCOTUS, and MIMIC datasets.  Performance is reported in F-scores $\times$ 100.}
\footnotesize
\begin{tabular}{l@{\hspace{6pt}}l@{\hspace{6pt}}c@{\hspace{4pt}}c@{\hspace{4pt}}c@{\hspace{4pt}}c@{\hspace{4pt}}c@{\hspace{4pt}}c@{\hspace{4pt}}c@{\hspace{4pt}}c@{\hspace{4pt}}}
\toprule 
\multirow{2}{2.5cm}{PTM} & \multirow{2}{2.5cm}{Model}
& \multicolumn{2}{c}{ECHR} 
& \multicolumn{2}{c}{SCOTUS}
& \multicolumn{2}{c}{EURLEX}
& \multicolumn{2}{c}{MIMIC}
\\
\cmidrule(lr){3-4}
\cmidrule(lr){5-6}
\cmidrule(lr){7-8}
\cmidrule(lr){9-10}
 &
& macro-F1 & $\mu$-F1 
& macro-F1 & $\mu$-F1
& macro-F1 & $\mu$-F1
& macro-F1 & $\mu$-F1 \\


\midrule
\multirow{2}{2.5cm}{BERT}  & Emb + MLP  & 36.32 & 55.56 & 40.6 &	61.79 & 25.86 & 51.44 & 49.06 & 62.99 \\
 & Emb$_{SimCSE}$ + MLP  & 48.69 &	61.37  & 45.50 & 61.14 & 33.43 & 57.07 & 54.59 & 67.13 \\
 & Emb$_{ESimCSE}$ + MLP & 50.83	
 & 64.41  & 52.12 & 64.07 & 34.16 & 54.56 & 57.05 & 68.08\\
 & Emb$_{CPE}$ + MLP & \textbf{54.77} & \textbf{66.98} & \textbf{54.56} & \textbf{66.78} & \textbf{40.96} & \textbf{63.68} & \textbf{57.12} & \textbf{68.20} \\
\midrule 
\multirow{2}{2.5cm}{RoBERTa} & Emb + MLP & 35.27 & 55.64 & 32.77	& 57.92 & 21.88
 & 35.05 & 49.63 & 64.27\\
 & Emb$_{SimCSE}$ + MLP  & 49.04  & 59.4 & 48.28 & 62.71 & 35.13 & 53.64 & 55.72	& 66.99\\
 & Emb$_{ESimCSE}$ + MLP  & 45.27  & 56.61 & 57.73 & 67.85 & 33.76 & 54.14 & 57.47 & 67.32 \\
 & Emb$_{CPE}$ + MLP & \textbf{56.02}  & \textbf{65.58} & \textbf{57.85} & \textbf{68.07} & \textbf{41.94} & \textbf{63.44} & \textbf{61.01} & \textbf{69.42} \\

\midrule
\multirow{2}{2.5cm}{LegalBERT} & Emb+ MLP  & 52.63  & 65.31 & 44.93 & 65.42 & 17.78 & 39.47 & \multicolumn{2}{l}{Not considered} \\
 & Emb$_{SimCSE}$ + MLP & 57.52	
  & 68.81 & 57.55 & 68.92 & 40.87 & 61.58 & \multicolumn{2}{l}{Not considered} \\
 & Emb$_{ESimCSE}$ + MLP  & 56.47 & 68.29 & 56.08 & 67.78 & 40.40
 & 62.59 & \multicolumn{2}{l}{Not considered} \\
 & Emb$_{CPE}$ + MLP & \textbf{57.74} & \textbf{69.39} & \textbf{59.45} & \textbf{71.85} & \textbf{42.16}
 & \textbf{64.17}
 & \multicolumn{2}{l}{Not considered} \\
 
 \midrule
\multirow{2}{2.5cm}{ClinicalBioBERT} & Emb + MLP & \multicolumn{6}{l} {not considered for legal documents} & 55.84 & 68.9\\
 & Emb$_{SimCSE}$ + MLP & \multicolumn{6}{l} {not considered for legal documents} & 62.74 & 71.06 \\
 & Emb$_{ESimCSE}$ + MLP & \multicolumn{6}{l} {not considered for legal documents} & 60.39 & 70.94 \\
 & Emb$_{CPE}$ + MLP & \multicolumn{6}{l} {not considered for legal documents} & \textbf{63.89} & \textbf{71.72} \\

\bottomrule
\end{tabular}

\label{table-classHT}
\end{table*}

\paragraph{Document Embedding Baseline Models} \label{sec:baseline}
We compare classifier model performance in the following settings.

\noindent\emph{Pre-trained models Embedding + MLP Classification:} In this setting, we obtained the document embeddings using the existing pre-trained parameters of the hierarchical pre-trained language models (BERT, RoBERTa, LegalBERT, ClinicalBioBERT), and the Longformer model. On top of these document embeddings, we applied a MLP classification layer to predict the legal and medical labels. During training, only the parameters of the MLP layers are updated, while the parameters of the pre-trained models are fixed.
    
\noindent\emph{SimCSE Embedding + MLP Classification:} 
     SimCSE~\cite{gao2021simcse} proposed a contrastive learning framework that employs dropout for data augmentation. We train the SimCSE framework using hierarchical pre-trained language models and the Longformer model on long documents. After producing document embedding from the SimCSE framework, we  apply an MLP layer for determining the legal and medical labels. 
     
\noindent\emph{ESimCSE Embedding + MLP Classification:} The Enhanced SimCSE~\cite{wu2022esimcse} contrastive learning framework applied word repetition operation to construct positive pair. We train the ESimCSE on long documents by employing hierarchical pre-trained language models and the Longformer model. We trained an MLP classifier on top of fixed ESimCSE document embedding.

\paragraph{End-to-end Finetuning Baseline Models} 
We compared our approach against the following state-of-the-art fine-tuning methods:

\noindent\emph{ Hi-LegalBERT~\cite{chalkidis2022lexglue}:} The hierarchical Legal-BERT encodes each text chunk via a pretrained Legal-BERT [CLS] embedding. These chunk embeddings are then fed into a two-layer Transformer to capture inter-chunk context, then max-pool the context-aware chunk vectors into a single document embedding for classification. 
\noindent\emph{LSG Attention~\cite{condevaux2023lsg}:} LSG architecture employs Local, Sparse, and Global attention. LSG is based on block
local attention for short-range context, structured sparse attention for mid-range dependencies, and global attention to improve information flow. LSG produces competitive results for classifying long documents.  

\noindent\emph{LegalLongformer~\cite{mamakas2022processing}:} LegalLongformer adapts the Longformer architecture by initializing all parameters from LegalBERT and then fine-tuning the entire model on downstream long document classification tasks.

\noindent\emph{HAT \cite{chalkidis2022exploration}:} (Hierarchical Attention Transformer) employs a two-stage encoder: a segment-wise transformer to capture local context, followed by a cross-segment transformer to model inter-segment interactions. This hierarchical approach enables efficient and accurate classification of long documents.

We propose generalized CPE which can be applied to any hierarchical attentional document such as ToBERT~\cite{chalkidis2021paragraph}, RoBERT~\cite{pappagari2019hierarchical}, or sparse attention models such BigBird~\cite{DBLP:conf/nips/ZaheerGDAAOPRWY20}. By leveraging more advanced methods, the CPE can achieve better performance. We output the performance of the CPE  using an advanced hierarchical attentional document encoder in the table~\ref{tab:varHT}.

\section{Evaluation and Results} \label{sec:EvalRes}
We conduct a comprehensive evaluation of the proposed CPE framework by benchmarking against a hierarchical transformer encoder, a sparse-attention Longformer model, and state-of-the-art hierarchical transformer variants. Experiments are carried out on standard long legal and medical datasets under frozen-embedding and full end-to-end fine-tuning settings.

\begin{table*}
\scriptsize
\centering
\caption{Classification model performance on document embedding produced by contrastively pre-trained sparse attention models on ECHR, SCOTUS, EURLEX, and MIMIC datasets.  Performance is reported in F-scores $\times$ 100.}
\begin{tabular}{llcccccccc}
\toprule 
\multirow{2}{1cm}{PTM} & \multirow{2}{1cm}{Model}
& \multicolumn{2}{c}{ECHR} 
& \multicolumn{2}{c}{SCOTUS}
& \multicolumn{2}{c}{EURLEX}
& \multicolumn{2}{c}{MIMIC}
\\
\cmidrule(lr){3-4}
\cmidrule(lr){5-6}
\cmidrule(lr){7-8}
\cmidrule(lr){9-10}
 &
& macro-F1 & $\mu$-F1 
& macro-F1 & $\mu$-F1
& macro-F1 & $\mu$-F1
& macro-F1 & $\mu$-F1 \\
\midrule
\multirow{2}{2.5cm}{Longformer}  & Emb + MLP  & 35.89 & 50.35 & 27.35
 & 50.28 & 25.83	 & 54.45 & 44.55 & 62.09 \\
 & Emb$_{SimCSE}$ + MLP  & 35.15 & 46.81 & 36.82 & 52.64 & 32.01 & 53.61 & 45.36 & 60.77 \\
 & Emb$_{ESimCSE}$ + MLP & 32.92	
 &	46.29  & 35.33 & 53.35 & 35.38 & 55.85 & 46.06 & 61.44 \\
 & Emb$_{CPE}$ + MLP & \textbf{48.94} & \textbf{59.71} & \textbf{47.46} & \textbf{62.57} & \textbf{43.24} & \textbf{64.57} & \textbf{53.06} & \textbf{64.93} \\
\bottomrule
\end{tabular}

\label{table-classLongformer}
\end{table*}

\paragraph{Evaluation of Hierarchical Representation}

Legal and medical topic classification results on fixed document representation via hierarchical transformer results in presented in Table~\ref{table-classHT}. From the table, we can observe that MLP classifiers with document embedding from various PTM produce worse performance on all datasets. It is evident that self-supervised contrastive learning SimCSE, ESimCSE, and CPE improves the classification performance of PTM document embedding across the datasets. The SimCSE embedding improves the performance of BERT embedding by 12\% in terms of the macro-F1 score on the ECHR. While improvement of ESimCSE over BERT embedding is 14\%. However, the CPE embedding achieves the best performance on all datasets using BERT PTM. Specifically, CPE improves macro-F1 scores approximately by 6\%, and 4\% on BERT embedding of SimCSE, ESimCSE for ECHR, dataset. Similar improvement can be observed on SCOTUS, EURLEX, and MIMIC.

The MLP classifiers based on Bert-base model perform poorly for both legal and biomedical datasets. The potential reason is that legal and biomedical terminology are not well represented in the generic corpora of the BERT model. So we also perform experiments with LegalBERT PTM, BERT version pre-trained on legal corpora. Similarly, we utilize ClinicalBioBERT which is pre-trained on medical documents. The MLP classifier leveraging LegalBERT and ClinicalBioBERT document embeddings demonstrates enhanced performance compared to the classifier based on BERT document embeddings. In-domain knowledge appears to be particularly crucial for the SCOTUS dataset, resulting in a 5\% improvement in macro F1 score of LegalBERT Embedding$_{CPE}$ MLP over BERT Embedding$_{CPE}$ MLP. In comparison, the improvements on the ECHR and EURLEX datasets over BERT Embedding$_{CPE}$ MLP models are 3\% and 2\%, respectively. This is because the SCOTUS dataset is more domain-specific in its language compared to the other datasets. Similarly, in the MIMIC dataset, ClinicalBioBERT Embedding$_{CPE}$ shows a notable 3\% enhancement in performance compared to BERT Embedding$_{CPE}$.

The MLP classifier utilizing RoBERTa embeddings shows better performance compared to BERT-base model and achieves comparable results to the classifiers utilizing LegalBERT and ClinicalBioBERT embeddings. This improved performance can be attributed to RoBERTa's pre-training on larger generic corpora and more extensive vocabulary. The proposed RoBERTa Embedding$_{CPE}$ MLP outperforms SimCSE, ESimCSE MLP models on ECHR, SCOTUS, EURLEX, and MIMIC. 
The proposed CPE outperforms the document Emb, Emb$_{SimCSE}$, Emb$_{ESimCSE}$ employing various PTM encoders across datasets. The CPE training produced better performance using LegalBERT and ClinicalBioBERT. These results suggest that CPE performance improves with domain-adapted PTM. Furthermore, results indicate the dropout augmentation or simple repetition of words to construct positive pairs and generate text embeddings may not yield significant improvements at the document or paragraph level embeddings.

\paragraph{Evaluation on Longformer}
Table~\ref{table-classLongformer} shows the MLP classifier results on document representation produced by the contrastively trained Longformer model. The Longformer + MLP classifier performs poorly in all datasets because the model fails to model global context and paragraph-level interaction. It is evident that self-supervised strategies of (SimCSE, ESimCSE, and CPE) improve the results in the Longformer models. The proposed CPE method has a clear advantage over SimCSE and ESimCSE methods, producing 12\% and 14\% better macro F1-score on the ECHR dataset and 10\% and 12\% improvement in terms of macro F1-score on the SCOTUS dataset. 
Similarly, on the MIMIC dataset, the performance gain from CPE is 8\% and 7\% in macro F1 over SimCSE and ESimCSE, respectively.
Based on the results, it is clear that hierarchical transformer models perform better than Longformer-based models on the ECHR, SCOTUS, and MIMIC datasets. The only dataset where Longformer models demonstrate superior performance is EURLEX. This improved performance on the EURLEX dataset can likely be attributed to its shorter length, which allows the Longformer model to capture all relevant information more efficiently than the hierarchical transformer model.
Overall, our self-contrastive CPE methods, which are based on Longformer and hierarchical transformer, yield better results compared to SimCSE and ESimCSE models. This demonstrates the effectiveness of our method. Stacking MLP classifiers on top of document embeddings proves to be an effective approach for encoding long documents.

\begin{table} 
\centering
\small
\caption{Performance comparison of end-to-end fine-tuned classification models on the ECHR and SCOTUS datasets. } 
\begin{tabular}{l@{\hspace{6pt}}c@{\hspace{4pt}}c@{\hspace{4pt}}c@{\hspace{4pt}}c@{\hspace{4pt}}}
\toprule
	\multirow{2}{2.5cm}{Model} & \multicolumn{2}{c}{ECHR} & \multicolumn{2}{c}{SCOTUS}      \\
\cmidrule(lr){2-3}
\cmidrule(lr){4-5}
& macro-F1 & $\mu$-F1  & macro-F1 & $\mu$-F1 \\
\midrule
Hi-LegalBERT & 64.0 & 70.0 & 66.5 & 76.4 \\
LSG & 60.3 & 71.0 & 63.7 & 73.3  \\
LegalLongformer  & 63.6 & 71.7 & 66.9 & 76.6\\
HAT  & - & 79.8 & 59.1 & 70.5 \\
Ours  & \textbf{66.1} & 72.6 & \textbf{67.3} & \textbf{77.5} \\
\bottomrule
\end{tabular}
\label{tab:FinetuneHT}
\end{table}

\paragraph{Evaluation on End-to-end Finetuning Setting}
To further solidify our findings, we evaluated our proposed CPE encoder in an end-to-end fine-tuning setting. We applied the LegalBERT CPE encoder to encode different chunks. Rather than using simple average pooling, we employed a two-layer transformer encoder to aggregate the information from different chunks. All parameters are trained in an end-to-end manner. 
The results for the classification of long documents are presented in Table~\ref{tab:FinetuneHT}. On the ECHR dataset, our model achieves a macro-F1 score of 66.1 an improvement of 2.5 points over LegalLongformer and 2.1 points over Hi-LegalBERT—and a micro-F1 score that is 0.9 points higher than LegalLongformer and 2.6 points higher than Hi-LegalBERT. Compared with LSG, we see even larger margins +5.8 pp macro and +1.6 pp micro on ECHR, +3.6 pp macro and +4.2 pp micro on SCOTUS.
HAT delivers a strong micro-F1 (79.8) on ECHR but underperforms on SCOTUS macro (59.1), indicating less balanced class handling.
Our consistent gains in both macro- and micro-F1 confirm that our CPE encoder generates high-quality chunk representations, simplifying the downstream classification of both frequent and rare legal classes during fine-tuning.

\paragraph{Supplementary Evaluation and Analysis} Additional evaluations in the appendix highlight the robustness of the proposed CPE framework. As shown in Table 3, end-to-end fine-tuning with a hierarchical transformer improves performance over baselines like Hi-LegalBERT and LegalLongformer, confirming CPE’s effectiveness. An ablation study on chunk size (Table 6) shows 128-token chunks offer the best performance, balancing context and relevance. On the short-document BIOSQ dataset (Table 7), CPE outperforms contrastive baselines, demonstrating its adaptability beyond long texts. Embedding visualizations and similarity metrics (Figure 3) further confirm that CPE produces semantically coherent and discriminative document representations. 

\section{Conclusion} \label{sec:Conc}

We proposed a novel self-supervised contrastive learning framework, Chunk Prediction Encoders (CPE), for generating long document representations in legal and biomedical domains. CPE captures intra-document and inter-document relationships by aligning text chunks within a document and distinguishing them from other documents. Our experiments on legal and biomedical classification tasks demonstrated significant improvements in macro F1 scores, outperforming baselines. Additionally, leveraging domain-specific pre-trained models like LegalBERT and ClinicalBioBERT within CPE further boosted performance, emphasizing the value of domain adaptation in document representation. Our findings highlight the potential of CPE for enhancing document understanding in legal and medical domain.

\section*{Limitations}

While CPE demonstrates strong performance in long document representation, certain aspects warrant further exploration. Its effectiveness across diverse domains beyond legal and medical texts remains to be fully assessed. Additionally, while the model effectively captures intra-document relationships, further refinements could enhance its adaptability to varying document structures. Future work could explore extensions to multilingual datasets and cross-domain applications.

\section*{Acknowledgments}
This work was supported by ANR-22-CE23-0002 ERIANA and ANR Chaire IA Responsable.

\bibliography{custom}

\appendix


\section{Experimental Setting}

\subsection{Datasets}
The different datasets we use are the following. Table \ref{table-Datastats} reports statistics. 

\emph{ECHR:} The European Court of Human Rights (ECHR)~\cite{chalkidis2021paragraph} comprises around 11K cases of alleged human rights violations. Each case contains a list of facts or events from the case description and the task is to predict whether specific human rights articles have been violated among 10 labels.

\emph{SCOTUS:} The Supreme Court of the United States (SCOTUS)~\cite{chalkidis2022lexglue} contains complex cases that are not well solved by the lower court. 
 SCOTUS is a single-label classification task, which predicts the court opinions from a choice of 14 available issues such as criminal procedure, and civil rights, among others. The training set consists of 5K and the validation and test set contain 1.4k cases. 

\emph{EURLEX}~\cite{chalkidis2022lexglue}: European Union (EU) legislation dataset contains 65K cases from the European Union portal \footnote{\url{http://eur-lex.europa.eu/}}. The task is to predict its EuroVoc labels such as ( economics, trade, and healthcare). The training set consists of 55K and the validation and test set contain 5k cases. The average document length is approximately 1400 tokens. EUR-LEX is a multi-label classification dataset containing 100 labels (concepts).

\emph{MIMIC-III:} Medical information mart for intensive care (MIMIC-III)~\cite{Johnson2016MIMICIIIAF} dataset comprises 40K discharge summaries from US hospitals, with each summary mapped to one or more (International classification of diseases, ninth revision) ICD-9 taxonomy labels. We utilized labels from the first level of the ICD-9 hierarchy.

\emph{BIOASQ}~\cite{tsatsaronis2015overview}: The BIOASQ dataset comprises biomedical articles sourced from PubMed. Each article is annotated with concepts from the Medical Subject Headings (MeSH) taxonomy. We used the first levels of the MeSH taxonomy. The dataset is divided into train and test categories.

\subsection{Configuration}
We use the AdamW optimizer with a learning rate of $2e-5$ and weight decay of $0.001$. The model is trained for 3 epochs for self-supervised SimCSE, ESimCSE, and CPE settings, while the MLP classifier model is trained for 20 epochs. The classifier uses a batch size of 16, while the self-contrastive learning module uses a batch size of 4. We apply Max Pooling to the chunk representations to aggregate information across the chunks. All models are trained using NVIDIA Quadro RTX 8000 48GB GPU.

\begin{table}[t] 
\footnotesize
\centering
\caption{Statistics of the long and short documents dataset - Average text token length and number of train and test samples for self-contrastive pre-training} \label{tab:ds_dist}
\begin{tabular}{lccc}
\toprule
Dataset & Train & Test & Avg. token length   \\
\midrule
ECHR & 9000  & 1000 & 2050 \\
SCOTUS & 5000  & 1400 & 8000  \\
MIMIC & 30000  & 10000 & 3200  \\
EUR-LEX & 55000 & 5000 & 1400\\
Bioasq & 80000 &  20000 & 300\\
\bottomrule
\hline
\end{tabular}
\label{table-Datastats}
\end{table}

\paragraph{Evaluation of Advanced Hierarchical Representation} \label{AdHiTrans}

In Section~\ref{CPE_contrastive}, we introduced a generalized CPE that can be used with any long document encoder. We hypothesise that the performance of CPE  improves when using an advanced document encoder. To test this, we conducted experiments using Transformer over BERT (ToBERT) and Recurrence over BERT (RoBERT). For classification tasks, we kept the parameters of BERT fixed (frozen), and only the Transformer and LSTM encoder with MLP layer were learned during training. The results in Table~\ref{tab:varHT} show indeed that ToBERT improved the performance of the generalized HBERT by 6\% in macro F1 and 1.5\% in $\mu$ F1-score. On the SCOTUS dataset, ToBERT achieved a performance gain of approximately 4\% in both macro and $\mu$ F1-scores.

\begin{table} 
\centering
\small
\caption{Performance evaluation of classification model using different hierarchical transformer CPE encoders applied on BERT for ECHR and SCOTUS datasets. } 
\begin{tabular}{l@{\hspace{6pt}}c@{\hspace{4pt}}c@{\hspace{4pt}}c@{\hspace{4pt}}c@{\hspace{4pt}}}
\toprule
	\multirow{2}{2.5cm}{Model} & \multicolumn{2}{c}{ECHR} & \multicolumn{2}{c}{SCOTUS}      \\
\cmidrule(lr){2-3}
\cmidrule(lr){4-5}

& macro-F1 & $\mu$-F1  & macro-F1 & $\mu$-F1 \\
\midrule
HBERT+MLP & 54.77 & 66.98 & 54.56 & 66.78  \\
RoBERT+MLP  & 54.99 & 67.05 & 58.34 & 69.64\\
ToBERT+MLP  & 60.28 & 68.46 & 58.79 & 70.07 \\
\bottomrule
\end{tabular}
\label{tab:varHT}
\end{table}

\section{Ablation Study} \label{sec:ablstudy}

We conducted an ablation study to evaluate the impact of different chunk sizes, visualize the quality of the embedding space, and examine the performance of the CPE framework on short documents. 

\begin{figure}[t!]
\includegraphics[width=0.47\textwidth]{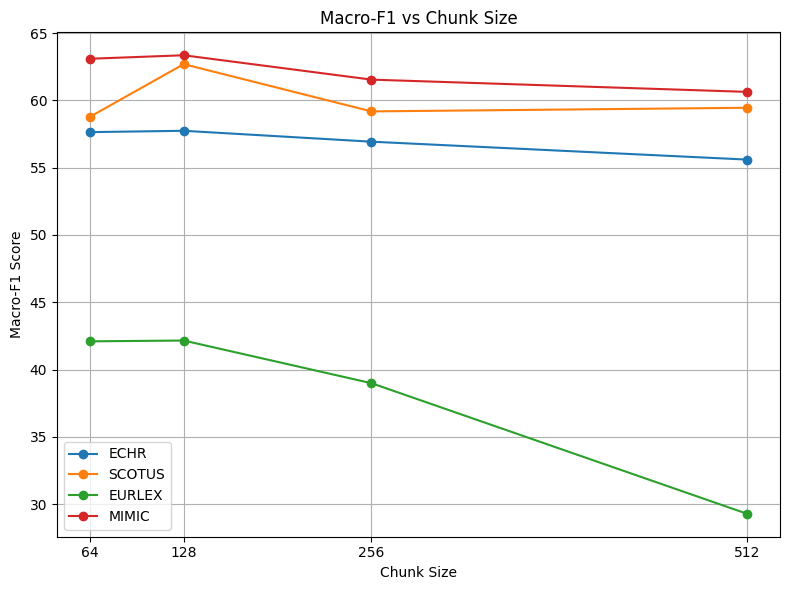}
\caption{Macro-F1 Scores vs. Chunk Size via hierarchical transformer encoders applied on LegalBERT for ECHR, SCOTUS, EURLEX, and ClinicalBioBERT for MIMIC datasets.}
\label{fig_chunk_macro-F1}
\end{figure}

\noindent\textbf{Impact of chunk length:}
Table~\ref{tab:chunkHT} and Figure~\ref{fig_chunk_macro-F1} summarize the CPE classification performance measured by macro‑F1 using a hierarchical Transformer encoder with chunk sizes of 64, 128, 256, and 512 across four datasets: ECHR, SCOTUS, EURLEX, and MIMIC. Fig~\ref{fig_chunk_macro-F1} illustrates how performance varies with chunk size. For the ECHR data set, on small chunk size of 64 produce 57.64 macro F1 score and remains stable for chunk size 128 and produce 57.74 micro F-1 score. The model performance then gradually decreases for large chunk sizes of 256 and 512. On the other hand, for SCOTUS dataset, there is a notable improvement when moving from a chunk size of 64 to 128, after which the performance slightly drops for larger sizes. The scores for MIMIC show a modest decline, on chunk size 64 to chunk size 512. This demonstrates relative robustness with only a slight decrease as the chunk size increases. Conversely, the EURLEX dataset exhibits its best performance at the smaller chunk sizes of 64 and 128, but shows a sharp decline at chunk sizes of 256 and 512. This suggests that EURLEX is highly sensitive to the chunk size parameter, likely because its shorter average text length means that larger chunks incorporate too much irrelevant detail.

\begin{table*}[t]
\centering
\footnotesize
\caption{Performance evaluation of classification model using different chunk size length via hierarchical transformer encoders applied on LegalBERT for ECHR, SCOTUS, EURLEX, and ClinicalBioBERT for MIMIC datasets. } 
\begin{tabular}{lcccccccc}
\toprule
	\multirow{2}{1cm}{Chunk Size} & \multicolumn{2}{c}{ECHR} & \multicolumn{2}{c}{SCOTUS} & \multicolumn{2}{c}{EURLEX} & \multicolumn{2}{c}{MIMIC}     \\
\cmidrule(lr){2-3}
\cmidrule(lr){4-5}
\cmidrule(lr){6-7}
\cmidrule(lr){8-9}

& macro-F1 & $\mu$-F1
& macro-F1 & $\mu$-F1
& macro-F1 & $\mu$-F1
& macro-F1 & $\mu$-F1 \\
\midrule
64 & 57.64 & 69.1 & 58.77 & 71.14 & 42.10 & 63.81 & 63.09 & 70.74 \\
128  & 57.74 & 69.39 & 62.69 & 73.29 &  42.16 & 64.17 & 63.35 & 70.79 \\
256  & 56.93 & 67.69   &  59.18 & 71.79 & 38.99	& 59.18 & 61.54 & 70.27 \\
512  & 55.60 & 67.29  & 59.45 & 71.85 & 29.29 & 43.63 & 60.63 & 68.90 \\
\bottomrule
\end{tabular}
\label{tab:chunkHT}
\end{table*}

\begin{table*}[t]
\footnotesize
\centering
\caption{Performance evaluation of classification model on short document BIOASQ dataset using BERT and ClinicalBioBERT encoders. } 
\begin{tabular}{l@{\hspace{6pt}}l@{\hspace{6pt}}c@{\hspace{4pt}}c@{\hspace{4pt}}c@{\hspace{4pt}}}
\toprule 
\multirow{2}{1cm}{PTM} 
& \multirow{2}{1cm}{Model}
& \multicolumn{2}{c}{Bioasq} \\
\cmidrule(lr){3-4}
& & \multicolumn{1}{c}{macro-F1} & \multicolumn{1}{c}{$\mu$-F1} \\
\midrule
\multirow{4}{2.5cm}{BERT} 
& Embedding + MLP             & 68.64 & 83.30 \\
& Embedding$_{SimCSE}$ + MLP   & 68.05 & 82.70 \\
& Embedding$_{ESimCSE}$ + MLP   & 68.04 & 82.66 \\
& Embedding$_{CPE}$ + MLP                           & \textbf{70.51} & \textbf{84.08} \\
\midrule
\multirow{4}{2.5cm}{ClinicalBioBERT}  
& Embedding + MLP                           & 68.05 & 83.60\\
& Embedding$_{SimCSE}$ + MLP   & 69.13 & 83.31 \\
& Embedding$_{ESimCSE}$ + MLP  & 68.77 & 83.17 \\
& Embedding$_{CPE}$ + MLP                    & \textbf{71.28} & \textbf{84.43}  \\
\bottomrule
\end{tabular}
\label{table-bio}
\end{table*}

\noindent\textbf{Performance on short document corpus}
To evaluate our CPE framework on short documents, we perform experiments on the BIOASQ dataset. We followed the method outlined in Section 3.2, but instead of using the Longformer encoder, we utilized ClinicalBioBERT and BERT features, setting the length of the positive chunk to 64 tokens. Table ~\ref{table-bio} reports classification results. The top rows show the performance of models using BERT embedding and the bottom rows display the performance of models using ClinicalBioBERT embedding. The ClinicalBioBERT Embedding$_{CPE}$ + MLP model produces the highest macro and micro F1 scores, achieving 71.28 and 84.43 macro F1 and micro F1-scores, respectively. This indicates that self-supervised CPE learning produces high-quality embeddings. Conversely, the state-of-the-art ClinicalBioBERT Emb$_{SimCSE}$ + MLP and ClinicalBioBERT Emb$_{ESimCSE}$ + MLP models does not enhance the performance of the baseline model Embedding + MLP. This suggests that using only dropout augmentation or basic word repetition to form positive pairs for generating text embeddings yields little benefit for document- or paragraph-level representations, even though these techniques perform very well for sentence embeddings. Results demonstrate that the proposed CPE method improves embedding derived from ClinicalBioBERT by around 4\% macro-F1.

\begin{figure*}[t]
  \includegraphics[width=0.48\linewidth]{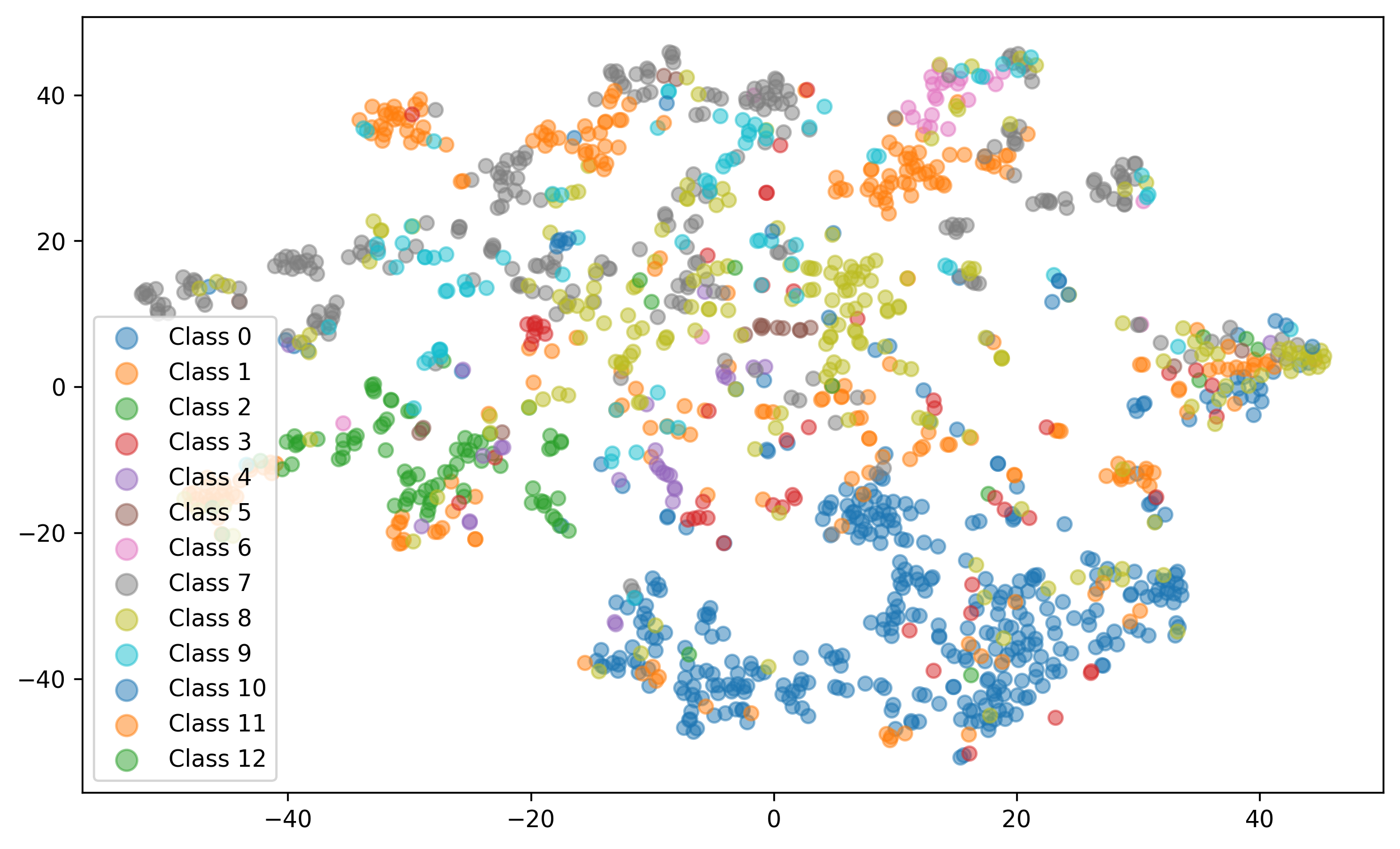} \hfill
  \includegraphics[width=0.48\linewidth]{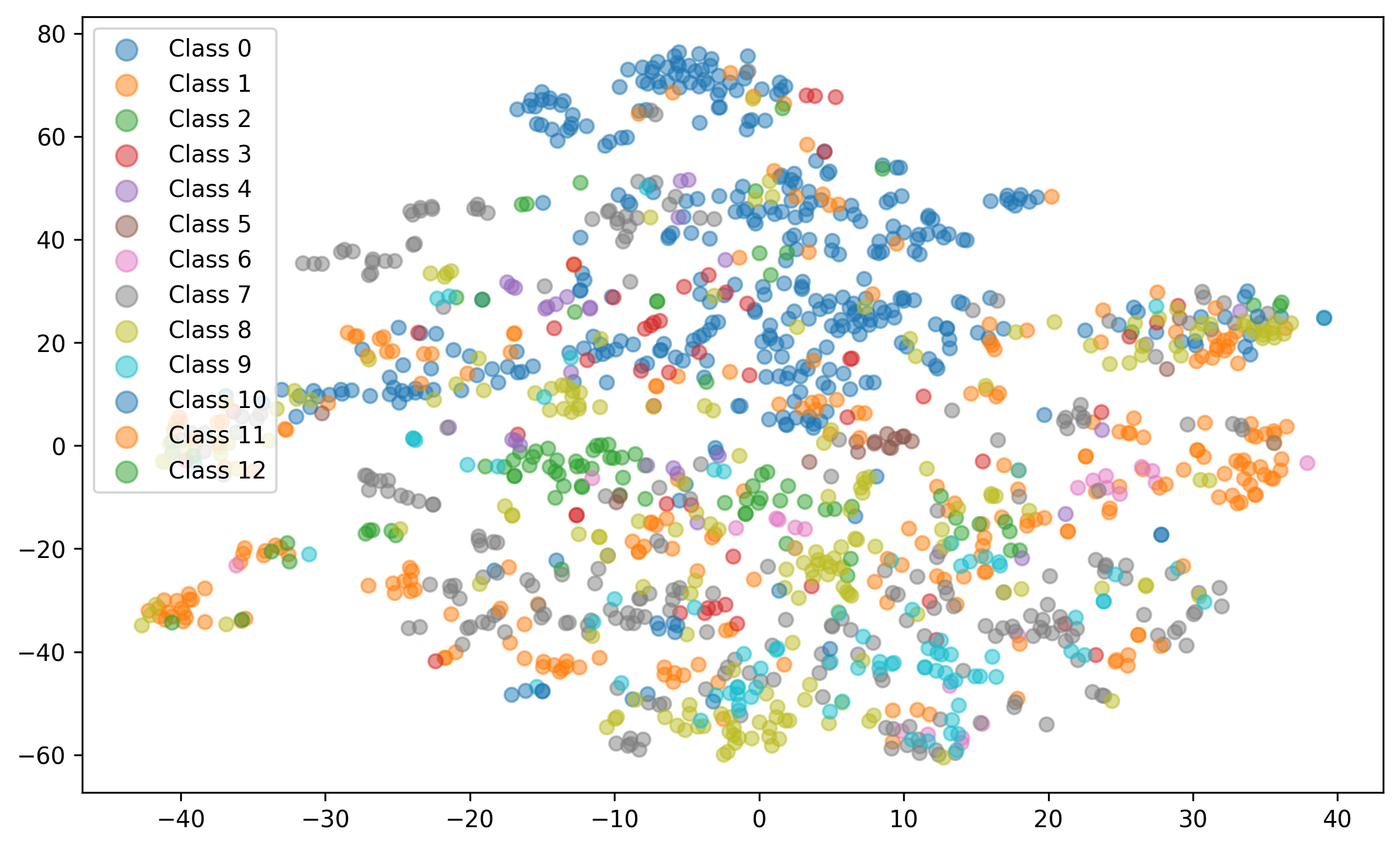}
  \caption {t-SNE visualization on the Scotus dataset with corresponding legal class.}
\label{fig_tsne_projection}
\end{figure*}

\noindent\textbf{Embeddings Quality}
Figure~\ref{fig_tsne_projection} shows the t-SNE projections of the CPE embeddings compared to the SimCSE baseline using the LegalBERT encoder on SCOTUS. As we can see, CPE demonstrates a higher quality of legal act encoding, as evidenced by more compact clusters. To quantify the comparison of visualized embeddings, we applied the DBScan clustering algorithm to the t-SNE projections. We evaluated clustering quality using completeness and homogeneity measures. As shown in Figure~\ref{fig_tsne_projection}, the completeness and homogeneity scores for CPE are 0.31 and 0.38, respectively, compared to 0.23 and 0.32 for SimCSE. This indicates a clear improvement in topic separation using the projected embeddings from CPE.

\noindent\textbf{Training Time}

Table \ref{tab:traintime} presents the training time required by each model on the ECHR and SCOTUS datasets.
Self-supervised contrastive learning using a CPE encoder requires approximately 3 h on ECHR and 1.5 h on SCOTUS dataset. In contrast, SimCSE and ESimCSE training takes place 4.5 on ECHR and 2.5 h on SCOTUS dataset.

SimCSE, and ESimCSE require significantly longer training times due to their document-level postive pair and negative pair. Our CPE training is more efficient because each positive and negative pair involves a single sampled chunk paired with an aggregated document context rather than encoding the entire document twice, reducing both computation and memory overhead. 

Furthermore, for the evaluation of the document embedding in the downstream with an MLP head (1.78M trainable parameters), the training is light, taking less than 10 minutes per epoch on our hardware.

\begin{table}[h]
\centering
\small
\caption{Training time (in hours or minutes) across different datasets.}
\label{tab:traintime}
\begin{tabular}{lcc}
\toprule
Model & ECHR (T) & SCOTUS (T) \\
\midrule
CPE         & 3 h   & 1.5 h      \\
SimCSE      & 4.5 h & 2.5 h     \\
ESimCSE     & 4.5 h   & 2.5 h   \\
\bottomrule
\end{tabular}
\end{table}

\noindent\textbf{Prompting LLAMA on long legal documents} 

Table~\ref{tab:prompt} demonstrates that the zero-shot prompting model LLAMA underperforms compared to embedding models (HBERT+MLP), primarily due to the extensive length of its prompts. Although the literature suggests that few-shot demonstrations can enhance model performance, the limited context window presents significant challenges when handling long documents, each averaging 4,096 tokens. Incorporating even one-shot examples into the prompt consumes nearly all available space, leaving insufficient room for the actual query. Furthermore, the LLAMA model tends to over-predict a limited number of classes while rarely predicting others, leading to imbalanced classification outcomes, as indicated by a low macro-F1 score.

\begin{table} 
\centering
\caption{Classification performance of the zero-shot LLAMA model on the ECHR and SCOTUS datasets.} 
\begin{tabular}{l@{\hspace{6pt}}c@{\hspace{4pt}}c@{\hspace{4pt}}c@{\hspace{4pt}}c@{\hspace{4pt}}}
\toprule
	\multirow{2}{2.5cm}{Model} & \multicolumn{2}{c}{ECHR} & \multicolumn{2}{c}{SCOTUS}      \\
\cmidrule(lr){2-3}
\cmidrule(lr){4-5}

& macro-F1 & $\mu$-F1  & macro-F1 & $\mu$-F1 \\
\midrule
HBERT+MLP & 54.77 & 66.98 & 54.56 & 66.78  \\
Llama3-8B-Instruct & 15.39  & 22.54 & 0.369 & 24.91  \\
\bottomrule
\end{tabular}
\label{tab:prompt}
\end{table}

\end{document}